\documentclass[letterpaper, 10 pt, conference]{ieeeconf}

\IEEEoverridecommandlockouts
\usepackage{cite}
\usepackage{amsmath,amssymb,amsfonts}
\usepackage{algorithmic}
\usepackage{graphicx}
\usepackage{textcomp}
\usepackage{subfigure}
\usepackage{xcolor}
\usepackage[bottom]{footmisc}

\title{AirTouch: Towards Safe Human-Robot Interaction Using Air Pressure Feedback and IR Mocap System}

\author{Viktor Rakhmatulin\textsuperscript{*}, Denis Grankin\textsuperscript{*}, Mikhail Konenkov\textsuperscript{*},  Sergei Davidenko\textsuperscript{*} \\ Daria Trinitatova, Oleg Sautenkov and Dzmitry Tsetserukou % <-this % stops a space
\thanks{The authors are with the Intelligent Space Robotics Laboratory, Center for Digital Engineering, Skolkovo Institute of Science and Technology (Skoltech), 121205 Moscow, Russian Federation. 
E-mail: {\tt\small \{viktor.rakhmatulin, denis.grankin, mikhail.konenkov, sergei.davidenko, daria.trinitatova, oleg.sautenkov, d.tsetserukou\}@skoltech.ru\}
}
}
\thanks{\textsuperscript{*}The authors are contributed equally to this work}
}

\begin{document}

\maketitle

\begin{abstract}
 The growing use of robots in urban environments has raised concerns about potential safety hazards, especially in public spaces where humans and robots may interact. In this paper, we present a system for safe human-robot interaction that combines an infrared (IR) camera with a wearable marker and airflow potential field. IR cameras enable real-time detection and tracking of humans in challenging environments, while controlled airflow creates a physical barrier that guides humans away from dangerous proximity to robots without the need for wearable devices. A preliminary experiment was conducted to measure the accuracy of the perception of safety barriers rendered by controlled air pressure. In a second experiment, we evaluated our approach in an imitation scenario of an interaction between an inattentive person and an autonomous robotic system. Experimental results show that the proposed system significantly improves a participant's ability to maintain a safe distance from the operating robot compared to trials without the system.
\end{abstract}

% \begin{IEEEkeywords}
% Human-robot collaboration, ...
% \end{IEEEkeywords}

\section{Introduction}  
% Structure
% 1) Intro: autonomous robots are penetrating our life and imposing risks on our life  
% 2) Safety standards for human-robot collaboration on the cobots example (SSM method described)
% 3) Haptic devices for HRC safety (to show the need for airpressure feedback)
% 4) Human tracking technologies in safety context (to underline the need for infracamera)? Expand ?
% 5) Novelty statement: novel approach  to ensure safe interaction with autonomous robots in Urban environment 

% General intro: autonomous robots are penetrating our life and imposing risks on our life  
The use of service robots in various industries has brought numerous benefits to businesses, such as increased efficiency and reduced labor costs. However, this deployment also poses potential risks to human safety and business operations due to the inattentiveness and forgetfulness of people \cite{bookOnSafety}. People tend to be careless and to neglect necessary safety precautions. This could result in incidents such as collisions, falls, or other forms of harm. To minimize these risks, it is essential to implement effective safety measures to promote responsible and cautious interaction between people and autonomous robots \cite{bookOnSafety2}. This study focuses on the safety aspects of human interaction with autonomous robots that charge electric vehicles, aiming to promote collision-free human-robot interaction (HRI).
% end of General intro  
 % Safety ensuring approaches from robot side from standards and papers 
 % Scalera et al. proposed the generation of stopping trajectories to regulate the robot's speed and avoid accidentsIn , Scalera et al. proposed the generation of stopping trajectories to regulate the robot's speed and avoid accidents . Palleschi et al. computed safety trajectories based on humans representation as points of interest \cite{human_repr_as_points}. Pereira \cite{pereira2017} developed models for online estimation of operator's workspace occupancy based on preliminary recorded data of human motion.

The International Organization for Standardization (ISO) provided the first comprehensive framework for safe Human-Robot Collaboration (HRC) in the technical specification 15066 \cite{ISO2016}. Speed and separation monitoring (SSM) is a method used to maintain a speed-dependent separation gap between the human and the robot throughout operation \cite{SSM}. This approach relies on intelligent robot speed control, as well as human motion modeling and tracking techniques. For instance, Zanchettin et al. utilized the trajectory scaling technique, which takes into account the current velocity of the robot and dynamically scales its trajectory to prevent collisions \cite{safetyRob2,SSMscalingAndDangerZones}. Another technique is to generate stopping trajectories in case of elevated risk of collision \cite{safetyStopping,human_repr_as_points}. However, most of the existing research on SSM has focused primarily on ensuring safety through the robot control system, with limited consideration given to the role of the operator, whose awareness of the intended actions of the robot is often limited.
%  end of safety ensuring approaches from robot side 
\begin{figure}[!t]
  \centering
  \includegraphics[width=0.98\linewidth]{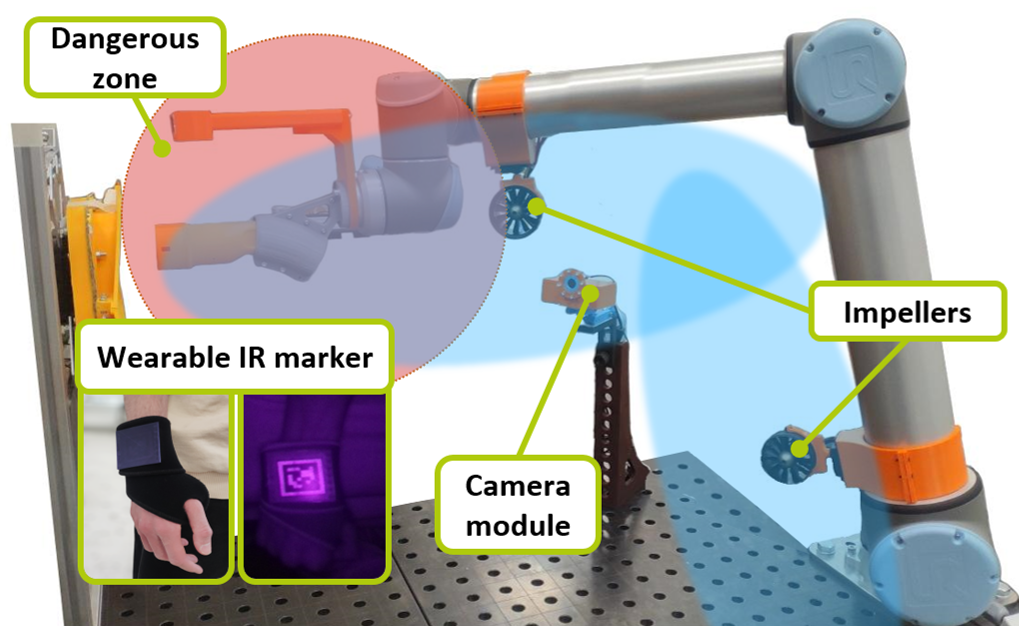}
  \caption{Conceptual diagram of a system for safe Human-Robot Collaboration in Urban Environment. The system includes impellers mounted on the robot to provide air-pressure feedback when the user intersects the dangerous zone (indicated in red). The effective working zones of the impellers are shown by the blue ellipses. The human's position is tracked using a wearable marker and an infrared camera.}
  \label{fig:mean_accuracy}
\end{figure}

\begin{figure*}[!t]
\vspace{0.3 cm}
  \centering
  \includegraphics[width=0.98\linewidth]{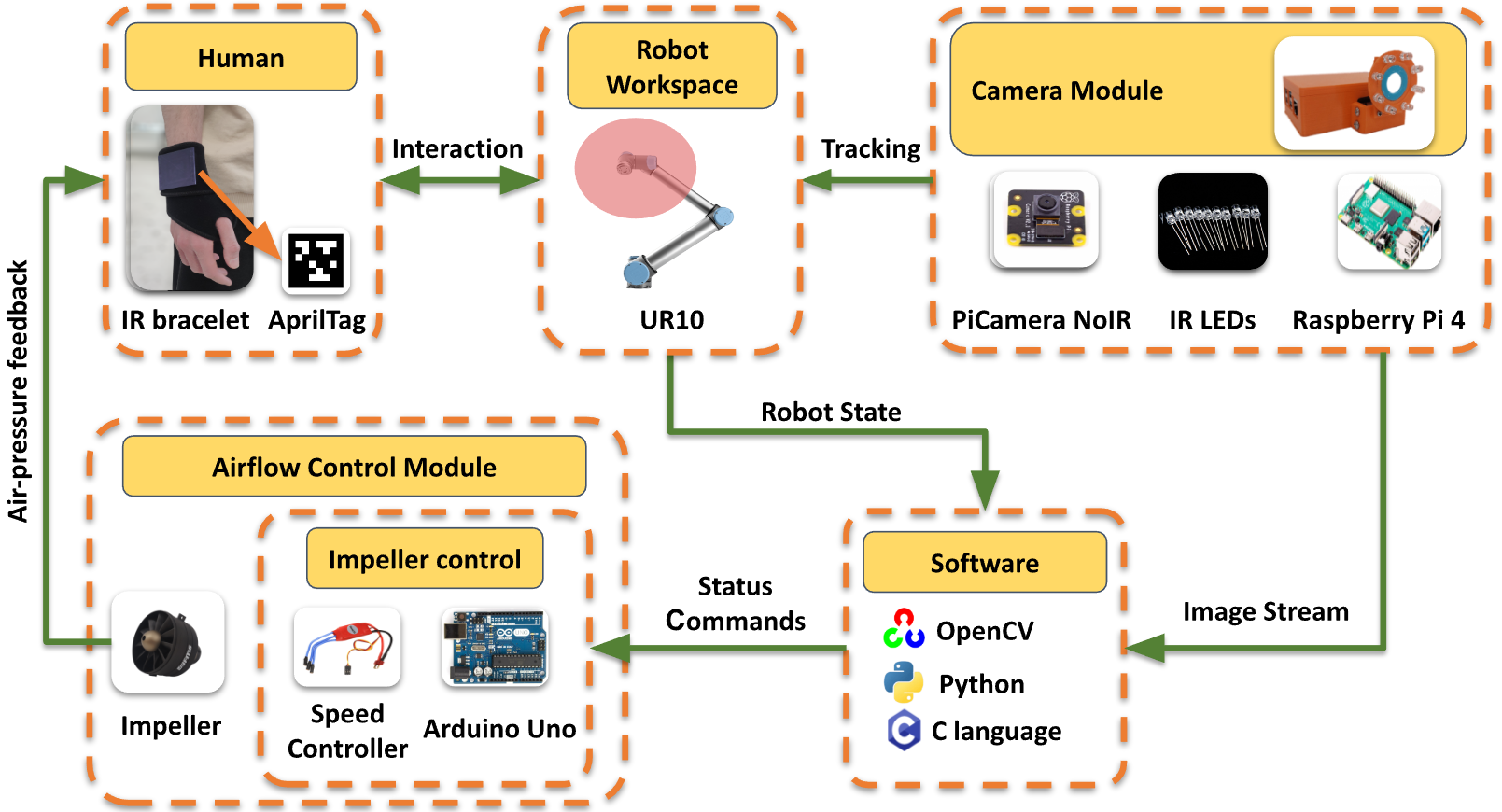}
  \caption{The architecture of the developed safety system for HRC.}
  \label{fig:architecture}
\end{figure*}

% Haptic devices for HRC safety (show that propeller is cool)
% minor paraphrase by chatgpt
Researchers have investigated the use of sensory augmentation to promote safe HRC. Vogel et al. \cite{Vogel2016} proposed a safety system that projects safety zones around a static robotic manipulator to provide visual feedback to the operator. Han et al. \cite{projecting_safety_zones} explored projections to visualize the intention of mobile robots towards pedestrians. However, a key limitation of these approaches is the need to track and interpret visual feedback, which can be challenging for inattentive individuals. On the contrary, wearable devices can be applied to provide efficient feedback on dangerous events and to facilitate spatial coordination of processes in a collaborative setting \cite{Ang2013}. Che et al. \cite{Che2018} developed a  vibration wristband to assist the operator avoid collisions with interfering mobile robots. Cabrera et al. \cite{cabrera2021cohaptics} proposed a wearable forearm device that provides skin stretch and vibrotactile feedback to increase operator awareness of the robot state during HRC. Although this approach demonstrated an improvement in decision-making towards safe HRC, wearable devices are not prevalent in urban environments. The same issue arises with sound feedback, as many people wear headphones and may not follow safety precautions.
% end of Haptic devices for safety

% Tracking in safety context 
To ensure safe HRC, human tracking plays a crucial role in addition to sensory feedback. Vogel et al. proposed a tactile floor to estimate the operator's foot position, but this method is limited as it cannot account for hand position and is not suitable for outdoor settings \cite{Vogel2016}. The frequency transceivers-based method localizes objects with good accuracy within the workspace \cite{rf_freq_local}. However, transceivers may be affected by interference from electronic devices such as phones and dynamic obstacles, making them challenging to use outdoors. InfraRed (IR) structured light cameras have been widely used in various HRC studies \cite{thermal_cam,human_repr_as_points} owing to their affordability and convenience. IR cameras provide distinct advantages for urban operations compared to visible spectrum cameras, primarily due to their efficient functioning in excessive and low lighting conditions, including nighttime scenarios. Costanzo et al. employed a thermal camera in conjunction with an IR depth sensor to improve accuracy of recognizing people.\cite{thermal_cam}. Therefore, designing and implementing safety systems for robots operating in urban environments requires careful consideration.
% end of tracking in safety context
  % and night vision, better visibility through obstructions like fog, smoke, and dust, 
% Statement of the inferred need for our system The proposed method offers accurate tracking and localization of humans, even in low-light or obscured conditions. 
% mig_cohaptics

This paper presents a novel approach for ensuring safe HRC in urban environments by utilizing a combination of an impeller and a motion capture (mocap) system based on a monocular IR camera with a wearable marker, inspired by Dogan et al. \cite{dogan2022}. Our approach is built upon the haptic potential field concept introduced in our previous study \cite{coboguider}, where we applied a wearable haptic device to render safety information and an \textit{Antilatency} mocap system for human position detection. In the present study, we utilize \textit{AprilTag} markers \cite{apriltag} made from IR reflective materials for efficient human tracking in complex urban environments. Besides, a monocular IR camera and wearable marker are cost-effective alternative  to structured light depth cameras. Furthermore, the marker can be embedded in ordinary clothing as a decorative element and carry supplementary information that facilitates human recognition for safety purposes. We suggest the use of airflow potential barriers as an alternative to wearable devices or sound and visual feedback to generate safety zones around autonomous robots. Our proposed approach has the potential to enhance the safety and efficiency of HRI in challenging urban environments.
% end of the Statement of the inferred need for our system

\section{System Architecture}
We designed a safety system consisting of two core elements: (1) a mocap system based on a monocular IR camera and a wearable marker, and (2) an airflow control system. The system architecture is shown in Fig. \ref{fig:architecture}.

\subsection{Hardware Description}
For detecting human position, we have developed a novel mocap system with IR visible \textit{AprilTag} markers \cite{apriltag}. The markers are printed on reflective tape and embedded under the 850 $nm$ IR filter, as described in \cite{dogan2022}. The markers are detected using an IR camera \textit{Raspberry Pi Camera Module 2 NoIR} equipped with eight 850 $nm$ IR LEDs. As a computing unit, we utilize \textit{Raspberry Pi 4 B} with $1\ Gb$ RAM to process the raw data from the camera. The time required to calculate a marker pose from a raw image is equal to 30 $\pm\ 2$ $ms$. The  camera localizes markers in changing lighting conditions and environmental factors typical for the outdoor setting. In addition, markers can be integrated into any clothing item while maintaining a natural appearance and functionality. For the current study, we designed a fabric wristband with a marker as an example of how the system can be naturally implemented into clothing.

% This approach allows us to give the user a clear and easily recognizable response to his actions. 
The air pressure feedback is provided by impeller \textit{FMS Ducted Fan Jet EDF Unit 11 blade} with the \textit{2840 KV3900} motor. The power of the airflow is controlled by a digital speed controller \textit{SimonK} with a continuous current rating of \textit{30A}. The speed controller is operated through an \textit{Arduino Uno} microcontroller board. 

We utilized a UR10 collaborative robot, which has a maximum reach of 1.3 $m$ and a payload capacity of up to 10 $kg$. The software for the system operates on a PC equipped with an Intel Core i7-10750H CPU with six cores and 12 threads, clocked at 2.6 GHz and 31.8 GB of RAM.

\subsection{Software description}
\begin{figure}[!t]
  \centering
  \includegraphics[width=0.95\linewidth]{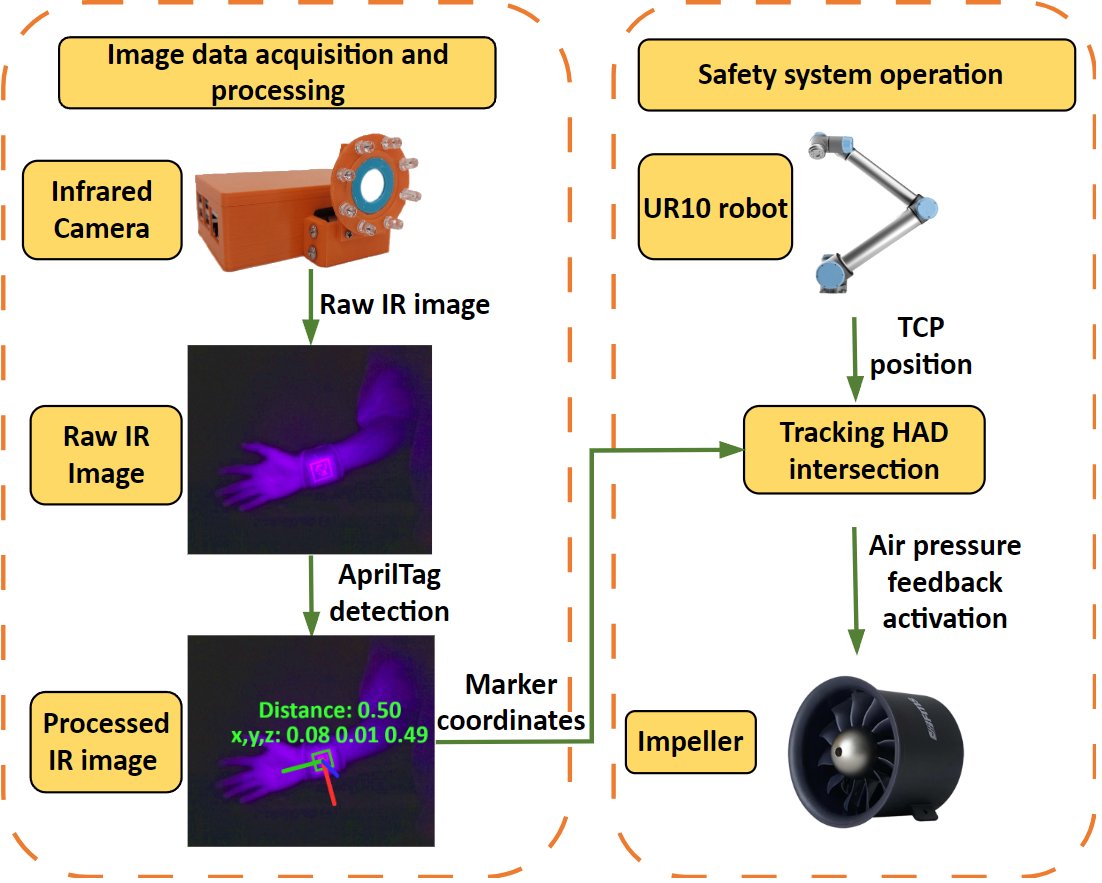}
  \caption{Illustration of algorithms for wearable marker tracking with an infrared camera and safety system operation. }
  \label{fig:algorithm_ill}
\end{figure}

Fig. \ref{fig:algorithm_ill} illustrates the working principle of the developed safety system. The impeller is activated when the distance between the marker and Tool Center Point (TCP) of the robot becomes below the threshold value, hereinafter referred to as haptic activation distance (HAD). We implemented the image data collection and processing, robot control, and communication with Arduino microcontroller using Python language. We utilized the multiprocessing standard library in Python\footnote{https://docs.python.org/3/library/multiprocessing.html} to minimize processing time per image and decrease delays of impeller activation. The code to control the impeller was written in the Arduino programming language.

% to prevent collisions between robots and humans
\section{User study I: Measurement of Safety Airflow Perception Accuracy.} 

\subsection{Experimental description}
\begin{figure}[!h]
%\vspace{0.3cm}
  \centering
  \includegraphics[width=0.95\linewidth]{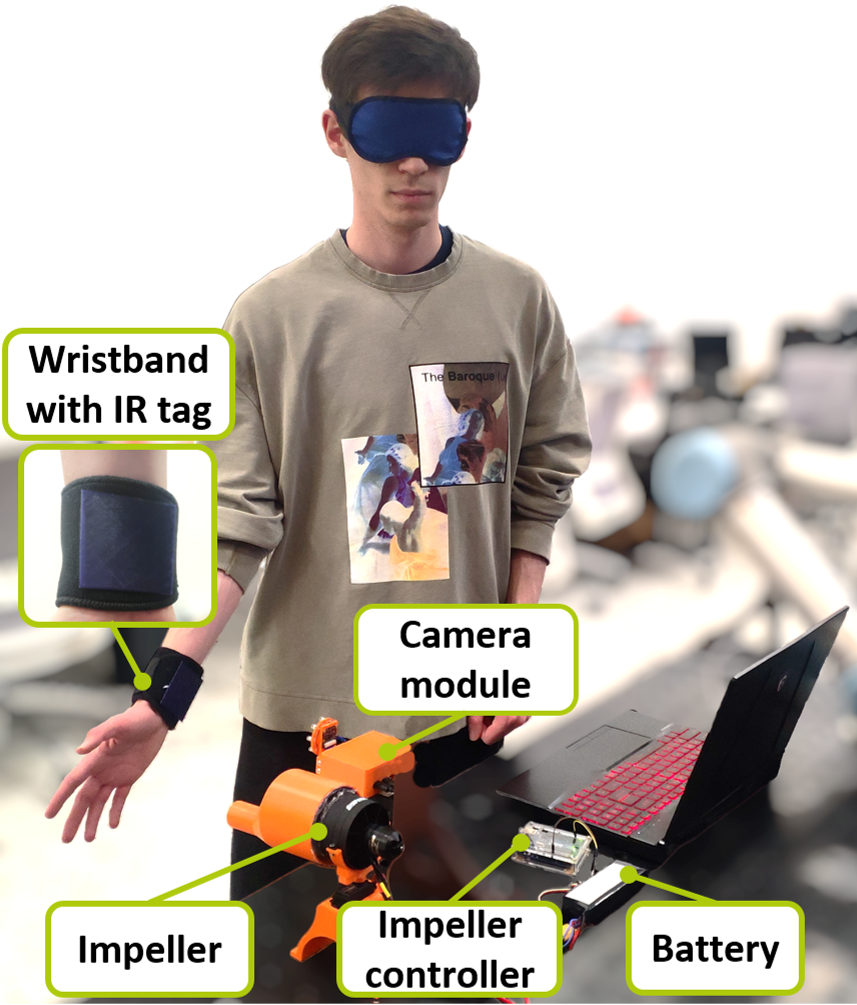}
  \caption{Experimental setup for measuring the accuracy of the air pressure perception.}
  \label{fig:exp1_setup}
\end{figure}

The main purpose of the first experiment was to evaluate the ability of users to perceive the airflow safety field. Ten right-handed participants (two females), volunteered to participate in the experiment, giving their informed consent. The experimental setup, as illustrated in Fig. \ref{fig:exp1_setup}, included an impeller controlled with a constant rotation regime and a camera module employed for tracking the participant's hand position.  

\subsection{Experimental procedure}
The task of the experiment for the participant was to identify two reference positions relative to the impeller: 0.25 and 0.35 $m$, relying on air pressure feedback. During the experiment, the participants were wearing a designed wristband with an IR marker and an opaque mask to eliminate visual feedback. Before the experiment, a training session was conducted in which each participant was instructed about the experimental procedure and tested the feeling of airflow from the impeller. For each condition (distances of 0.25 and 0.35 $m$), the participant's hand was moved to the reference position, and they were instructed to remember the airflow acting on it. The participant was then randomly relocated within the experimental setup to eliminate the impact of muscle memory. After that, the participant was asked to locate the hand to the reference position and report when the target position was reached. In total, each participant completed twenty attempts, ten for each reference position. For each participant, we measured the accuracy of reference position detection.

\subsection{Experimental Results and Discussion}
The accuracy of reference distance recognition by an individual participant is shown in Fig. \ref{fig:accuracy_individual}. We can notice that participants tended to overestimate the reference distance of 0.35 $m$, while the opposite was observed for 0.25 $m$. 

\begin{figure}[!h]
\begin{center}
\subfigure[Distance recognition accuracy of 0.25 $m$ for an individual participant.]{
\includegraphics[width=0.97\linewidth]{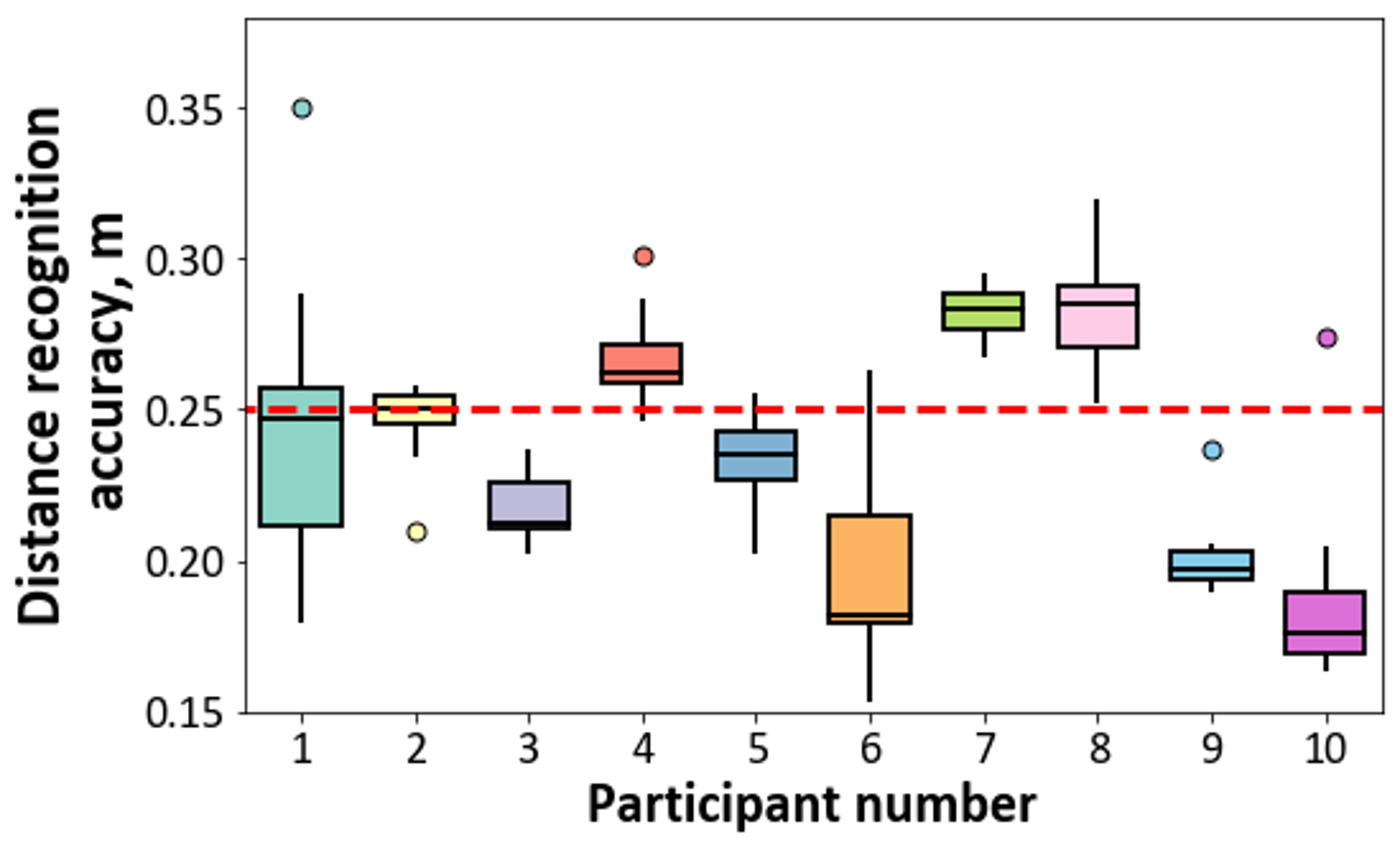}}
\subfigure[Distance recognition accuracy of 0.35 $m$ for an individual participant.]{
\includegraphics[width=0.97\linewidth]{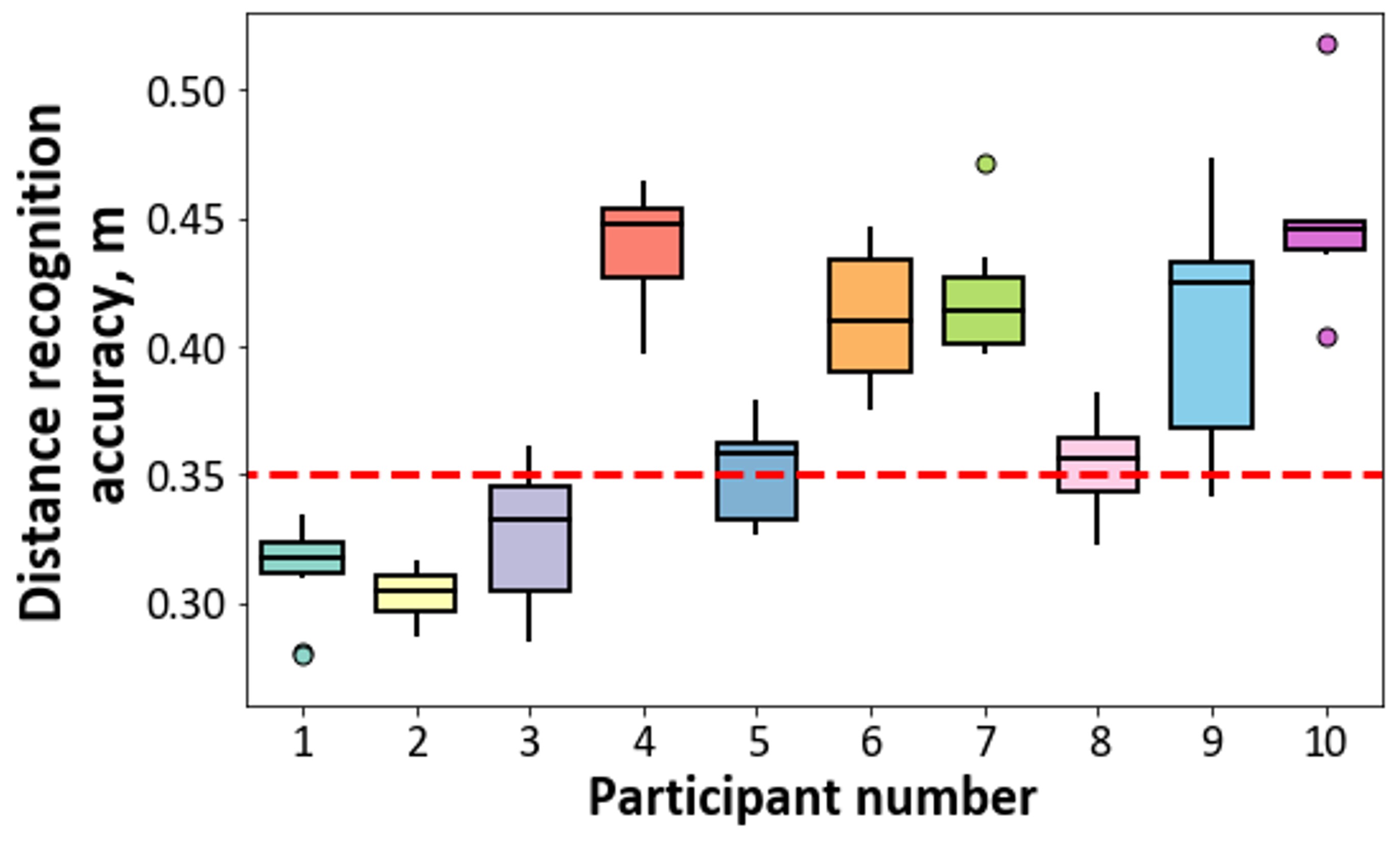}}
\caption{Distribution of participants' recognition accuracy of airflow generated for two reference distances of 0.25 $m$ and 0.35 $m$. The red dashed line represents the reference distance. }\label{fig:accuracy_individual}
\end{center}
\end{figure}

Fig.\ref{fig:mean_accuracy} shows the averaged participants' accuracy of distance recognition. To check the assumption of a normal distribution of the data, a Shapiro-Wilk test \cite{shaphiro} was performed which did not show the evidence to reject the hypothesis of normally distributed data for both reference positions ($p = .58$ and $p = .14$, respectively).
To evaluate the statistical significance of differences between measured reference positions, we applied a Paired Samples t-test, with a chosen significance level of $\alpha< 0.05$. According to the t-test results, there is a statistically significant difference between the distances of 0.25 and 0.35 $m$ ($T = -6.37,\ p < .001$). 
With the obtained results, we can conclude that the participants could confidently distinguish two close reference positions for the selected power mode of the impeller. The absolute mean error for a distance of 0.25 $m$ is $0.035 \pm 0.025\ m$, and for 0.35 $m$ is $0.051 \pm 0.035\ m$, which indicates that the air pressure generated by the impeller is sufficient to render the safety field for the participants.

\begin{figure}[!h]
%\vspace{0.3cm}
  \centering
  \includegraphics[width=0.96\linewidth]{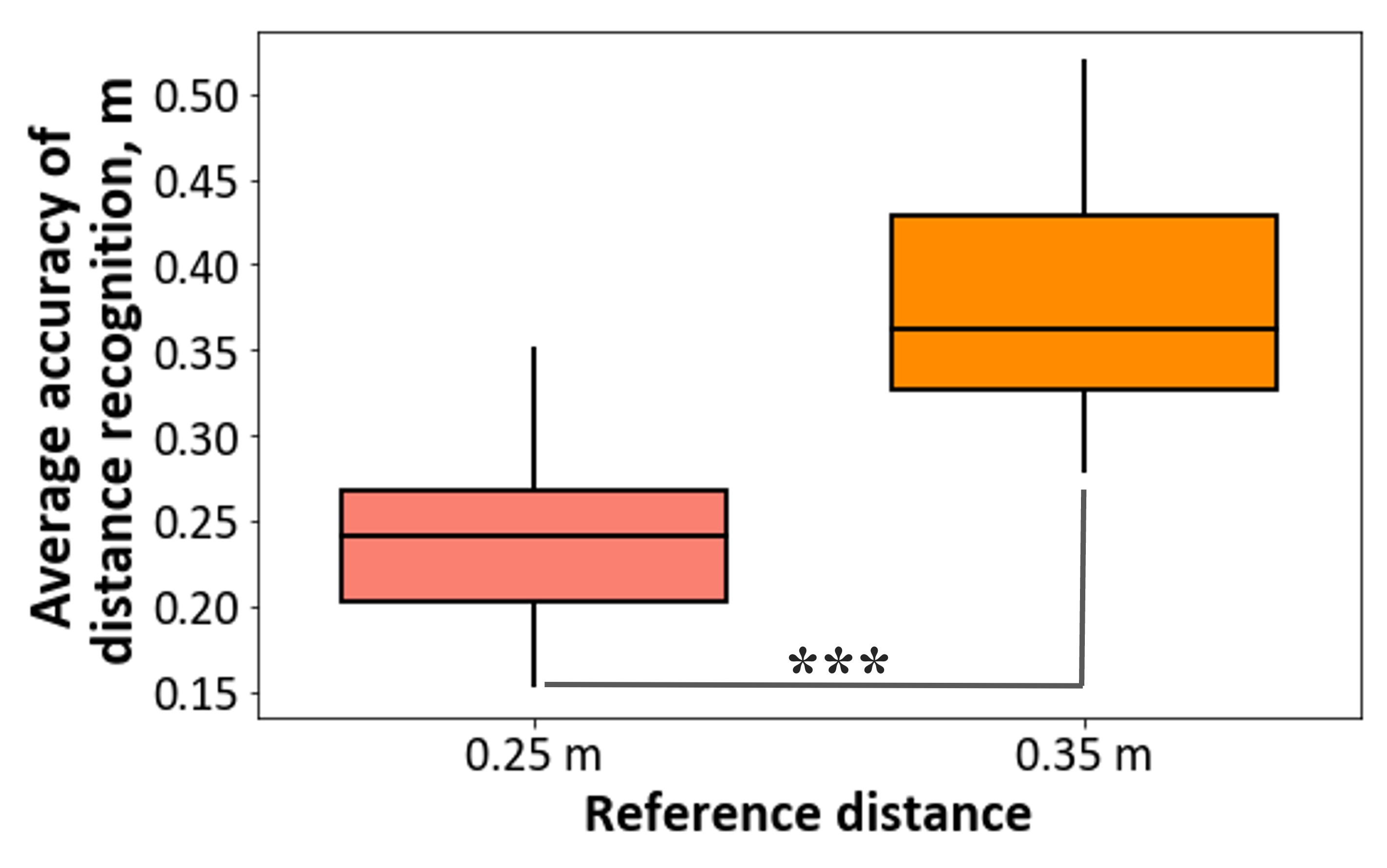}
  \caption{Average accuracy of distance recognition across all participants. An asterisk displays the statistical significance of the results ($^{***}:p \leq .001$).}
  \label{fig:mean_accuracy}
\end{figure}

\section{User Study II: Safe Interaction With Autonomous Charging Robot} \label{chapter:exp}

\subsection{Experimental description}
\begin{figure}[!h]
%\vspace{0.3cm}
  \centering
  \includegraphics[width=0.96\linewidth]{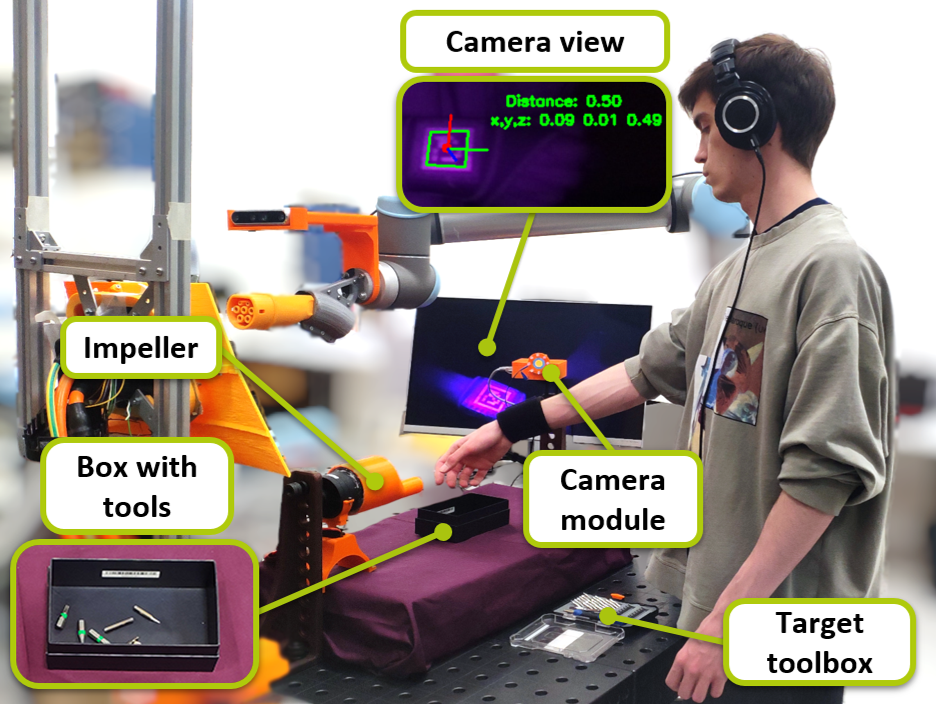}
  \caption{Experiment 2 overview: Human-robot interaction in a shared workspace. The participant collects the screwdriver bits, while the robot imitates the plug-in operation for the electric vehicle charging procedure.}
  \label{fig:exp2_overview}
\end{figure}

In this experiment, we aimed to assess the impact of a safety system design on maintaining a protective distance during a realistic scenario of collaboration with an autonomous charging robot. We imitated a case when an inattentive person loses the items from the bag nearby the robot and unintentionally intersect the trajectory of the operating robot. We designed an experimental setup with the collaborative robot UR10, mocap, and airflow systems as shown in Fig \ref{fig:exp2_overview}. The safety distance to the robot is set to be 0.35 $m$ and dangerous proximity distance is defined to be 0.25 $m$ as in our previous work \cite{coboguider}. Ten right-handed participants (2 females) volunteered to complete the study. The participants were informed about the experiment and agreed to the consent form. 

\subsection{Experimental procedure}
The experimental task for the participants was to collect the screwdriver bits of appropriate diameter into the toolbox while being in the workspace with a charging robot. In this experiment, each participant carried out the task in the presence of two types of feedback: Visual feedback only (V) and Visual and Air pressure feedback (VA). The air pressure feedback was used to inform the participant about the dangerous proximity and movements of the robot. We set the HAD equal to 0.35 $m$ and configured the safety system to activate the impeller upon crossing this distance. Before the experiment, a training session was conducted in which we demonstrated HAD to the participants and ensured that the marker was adjusted to enable stable detection.
During the experiment, the participants wore headphones with white noise to exclude sound feedback from the robot and safety system. Besides, the participants were instructed to track the position of a marker on a screen, which was intended to simulate a distracting environment. Each participant conducted two attempts to collect the screwdriver bits, one for each feedback condition. 

\subsection{Experimental Results and Discussion}
During the experiment, we measured the participant's distance to the robot TCP. Fig. \ref{fig:accuracy_individual} shows the example of the distance profiles for the 2 participants during the experiment. 
\begin{figure}[!h]
\vspace{0.3cm}
\begin{center}
\subfigure[Distance change profile with visual feedback only.]{
\includegraphics[width=0.9\linewidth]{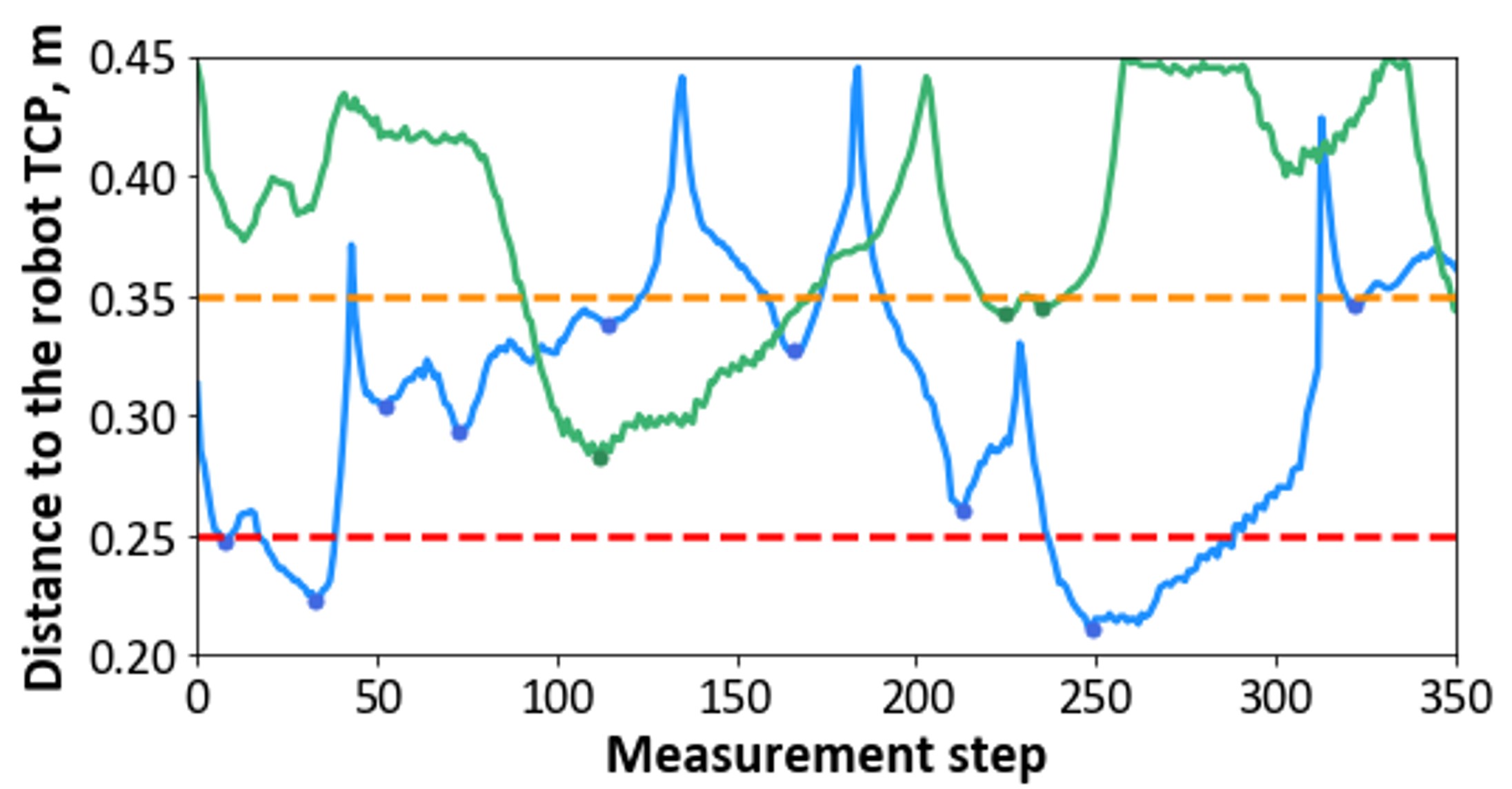}}
\subfigure[Distance change profile with visual and haptic (air pressure) feedback.]{
\includegraphics[width=0.9\linewidth]{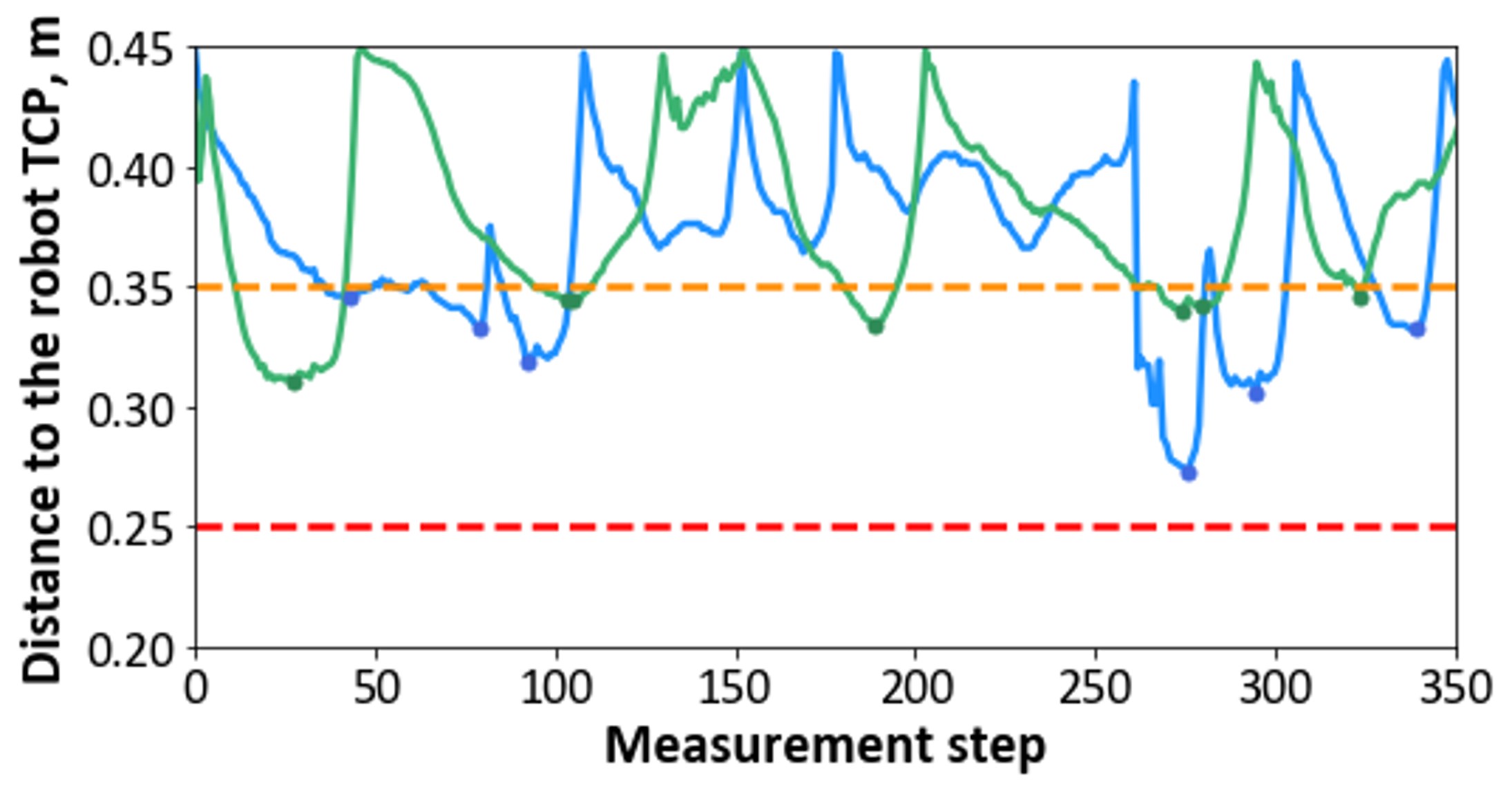}}
\caption{An example of the distance changes for the two participants during the experiment, with and without a developed safety system.  The orange and red dashed lines represent the haptic activation and dangerous proximity distances. }\label{fig:accuracy_individual}
\end{center}
\end{figure}

\begin{figure}[!h]
  \centering
  \includegraphics[width=0.95\linewidth]{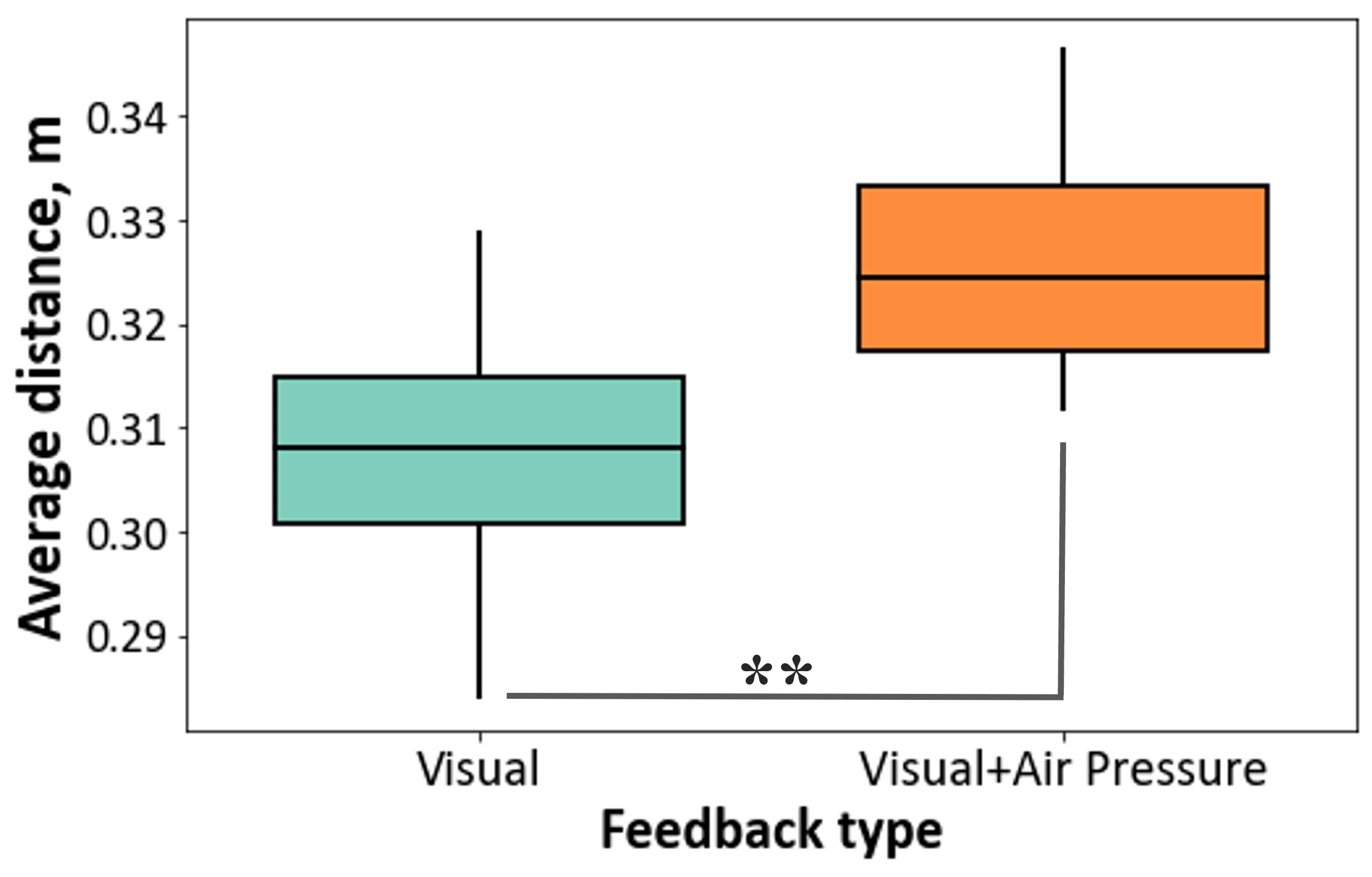}
  \caption{
  %The average participants' distances to the robot TCP inside the dangerous proximity zone for two feedback conditions.  
  A comparison of participants' average distances to the robot TCP in the dangerous proximity zone across two feedback conditions. An asterisk displays the statistical significance of the results ($^{**}:p \leq .01$).}
  \label{fig:mean_dist}
\end{figure}

To conduct a statistical evaluation of the obtained data, we discarded the distance measurements above the HAD for each participant and averaged the obtained results. Fig. \ref{fig:mean_dist} shows the relative distance between the participants' hand and the robot TCP for two experimental conditions. The Shapiro-Wilk test was applied to check the normality of data distribution, which did not show the evidence of non-normality ($p = .75$ and $p = .72$ for V and VA conditions, respectively). The average distance for V is $0.307 \pm 0.014 $ $m$, while for VA is $0.326 \pm 0.01$ $m$. To evaluate the statistically significant differences between the average distances for two feedback conditions, the results were analyzed using the Paired Sample t-test, with a chosen significance level of $\alpha < .05$. According to the t-test results, the average participants' distance to the robot TCP was statistically significant higher while using the developed safety system (VA condition) compared to pure visual feedback mode ($T = -3.52,\ p = .006$).

\section{Conclusion and Future Work}
In this study, we have proposed a novel approach to enhance the safety of HRI in an urban environment. Our approach comprises an air pressure feedback system and an IR tracking system based on a monocular camera and AprilTag wearable marker. Firstly, we conducted an experiment to measure the accuracy with which individuals perceive a safety distance rendered with airflow. According to the obtained results, we can conclude that participants can accurately perceive the safety field generated by the airflow safety system. Specifically, the absolute mean error at a distance of 0.25 $m$ was found to be $0.035 \pm 0.025\ m$, while the average error at a distance of 0.35 $m$ was $0.051 \pm 0.035\ m$. Subsequently, we evaluated our approach in a simulated interaction scenario of an inattentive person and an autonomous robotic charger for electric vehicles in a cross-condition comparison (visual feedback only vs. visual + air pressure feedback). The experimental results revealed that the use of air pressure feedback statistically significantly increased the average distance to the robot operating in dangerous proximity from $0.307\ m$ to 0.326 $m$.   
% We simulated a case when an inattentive person loses the items from the bag nearby the robot and unintentionally intersect the trajectory of the operating robot.

For future work, we are going to implement the simultaneous control of multiple impellers to improve rendering of the airflow safety field. Additionally, we plan to implement whole-body tracking with multiple markers and IR cameras. The technology has the potential to  enhance safety measures for human-robot interactions with autonomous robots in the urban environment.

\section*{Acknowledgment}
The authors would like to thank Miguel Altamirano Cabrera, PhD student at Skolkovo Institute of Science and Technology, for valuable support of the project.

\bibliographystyle{IEEEtran}
\bibliography{bib}

\end{document}